\documentclass{article}
\usepackage{spconf,amsmath,graphicx,amssymb}
\usepackage{hyperref}
\hypersetup{
    pdftitle={A new data augmentation method for intent classification enhancement and its application on spoken conversation datasets},
    }
\usepackage{cite}
\usepackage{booktabs,url}
\usepackage[table,xcdraw]{xcolor}

\def\T{{\cal T}}

\title{A new data augmentation method for intent classification enhancement and its application on spoken conversation datasets}
\name{\begin{tabular}{c} Zvi Kons, Aharon Satt, Hong-Kwang Kuo, Samuel Thomas,
\\Boaz Carmeli, Ron Hoory, Brian Kingsbury\end{tabular}}
\address{IBM Research AI}
\begin{document}
\ninept
\maketitle
\begin{abstract}
Intent classifiers are vital to the successful operation of virtual agent systems. This is especially so in voice activated systems where the data can be noisy with many ambiguous directions for user intents. Before operation begins, these classifiers are generally lacking in real-world training data. Active learning is a common approach used to help label large amounts of collected user input. However, this approach requires many hours of manual labeling work. We present the Nearest Neighbors Scores Improvement (\textbf{NNSI}) algorithm for automatic data selection and labeling.  The NNSI reduces the need for manual labeling by automatically selecting highly-ambiguous samples and labeling them with high accuracy. This is done by integrating the classifier's output from a semantically similar group of text samples. The labeled samples can then be added to the training set to improve the accuracy of the classifier.
We demonstrated the use of NNSI  on two large-scale, real-life voice conversation systems. Evaluation of our results showed that our method was able to select and label useful samples with high accuracy. Adding these new samples to the training data significantly improved the classifiers and reduced error rates by up to 10\%.
\end{abstract}
\section{Introduction}
\label{sec:intro}

One of the main components of any successful virtual assistant is the intent classifier. This classifier serves to identify a user's primary intention and direct him to the appropriate conversation path that will allow him to achieve his goal.
Generally, the training of these classifiers requires the system developer to identify all possible intents and provide text examples that match each one. This is a challenging task that requires the developer to predict good examples for possible users inputs. Once the system is running, the developers have to continuously monitor the conversations, identify problematic cases, and generate new labeled training examples to solve them.

In our work we address the case in which we are able to collect large amounts of user input from the conversation of a running virtual agent system. In common scenarios, the original training set is put together before operation begins and will typically not have enough examples to cover all the possible user inputs. In other words, the classifier is generally not trained to handle all scenarios before operation begins. To maintain and improve the system, the developer must go over the conversations, identify those with users inputs that were classified incorrectly, and then upgrade the classifier by adding these examples to the training set with the correct labels. When the number of available conversations is large, this could translate into many long hours of manual work by subject matter experts.

The same labelling problem applies to other scenarios in which we can produce a large set of unlabeled text samples. For example, one can create new samples by applying language models to generate variations of an existing set \cite{lambada}. 

Voice-based conversational systems introduce an additional  layer of complexity because the text to be classified is generated by a Speech-to-Text (STT) module. Often the text produced by the STT differs significantly from what the developer expected, because of STT errors and the differences between spoken and written text \cite{Wallace_Deborah,Gisela}.

Using active learning is a common approach to solving the labeling problem \cite{benvie_wayne_arnold1}. This method takes a set of unlabeled samples that was extracted from the collected conversations. It then uses a baseline classifier to generate labels for common phrases. Based on the frequency of the phrase and the confidence level of the classifier, the system developer can choose which samples to examine and then verify the classifier's label recommendations. This process can help reduce the amount of manual work, but its effectiveness is highly dependent on the accuracy of the baseline classifier.

In this paper, we suggest a different approach for handling a large set of unlabelled samples. We propose a weakly-supervised learning method that applies a baseline classifier on the unlabelled set. It then picks a subset of the samples and labels it with high accuracy. This new set of samples and labels can be added to the baseline classifier's training data to train a new classifier with improved accuracy.

The first challenge is to identify which samples in the unlabeled set are more likely to improve the classifier. We can divide the samples into two general groups:

\begin{itemize}
    \item \textbf{Low-ambiguity samples}: In these samples, the confidence score for one of the labels is much higher than all the others. This typically means that the classifier is confident about its results and these samples are likely to be labelled correctly.
    
    \item \textbf{High-ambiguity samples}: In these samples, the difference between the top label and at least one of the others is small. These samples are less likely to have a correct labeling.
\end{itemize}

We can choose to use the data that is more accurate by selecting the low-ambiguity samples and adding them to the training set. However, since the classifier is already confident about these, the benefit of adding them to the training data is likely to be small. On the other hand, adding high-ambiguity samples to the training only provides a larger benefit if we use the correct labels. Our second challenge is therefore to refine the labels that are generated by the classifier and improve their accuracy.

We address this problem by presenting the Nearest Neighbors Scores Improvement (\textbf{NNSI}) algorithm. This is a novel method that improves the labeling of the high-ambiguity samples by combining information from similar samples. We apply this procedure on the high-ambiguity samples and select those for which we can generate labels with lower ambiguity.

The main idea of NNSI is somewhat similar to other transductive weakly-supervised learning methods \cite{vanEngelen2020,10.1093/nsr/nwx106}. Those also use graphs constructed on the data to transfer labels from labeled edges to unlabeled ones. The accuracy strongly depends on the construction of the graphs, which may be as simple as $k$-NN or more complex ones \cite{learningwithgraphs,Lin_Xu_Zhang_2020}. The main differences in our NSSI method are that we propagate additional information along the graph and not just the labels. The NNSI also propagates information from all the nodes, and not just the labeled ones. However, we apply this information only to a subset of the unlabeled nodes.

Other weakly-supervised methods are different from our method because they require modification to the classifiers, the training algorithms, or the weights given to different samples \cite{TransductiveSVM,WeaklyLabeledSVMs}. 

In the following Section \ref{sec:nnsi} we provide a detailed description of the NNSI algorithm. Then, in Section \ref{sec:Experiments} we demonstrate how it can improve the accuracy of the intent classifier for two real-life datasets.

\section{NNSI algorithm}
\label{sec:nnsi}
We start with a baseline labeled set that holds pairs of text examples and intent labels $( T_k, L_k )$ with $T_k \in \T_L$ and $L_k \in \{ 1, \ldots, M \}$, and an unlabeled text set $\T_U$. The labeled set is used to train a baseline text classifier $C^0: \T \mapsto \mathbb{R}^M$. For any given text sample $T$, the classifier outputs a vector $S$ with a confidence-level score for each one of the possible $M$ class labels:
\begin{equation}
    C^0[T] = S = [ s_1, \ldots, s_M  ]
\end{equation}
with the goal $L_k = \mathrm{argmax}(C^0[T_k])$.

To help identify the samples for which the classifier is confident about its decisions, we define an ambiguity measure. We chose our score-ambiguity function $\Delta$ to be the difference between the highest and the second-highest confidence scores in the vector:
\begin{equation}
\Delta( S ) =  s_{(M)}  - s_{(M-1)}
\end{equation}
where $s_{(k)}$ is the order statistics of vector $S$, i.e., the $k$-th smallest value.

Our goal is to add more high-ambiguity samples (small $\Delta$ values) to the baseline training set. This is done by providing more accurate labeling for the high-ambiguity samples or by selecting the high-ambiguity samples whose labels are more accurate. Since the classifier output for one sample might not be accurate, we gather additional information from similar samples by relying on the smoothness assumption \cite{vanEngelen2020}.
We do this by looking at the classifier scores of similar text samples and combining them to generate more coherent results.
We evaluate the similarity of the text samples using a distance function $D( T_n, T_m )$ that estimates the semantic distance between two text samples.

The labeling process proceeds as follows. We start with an unlabeled, high-ambiguity text sample $T \in \T_U$, i.e.,
\begin{equation}
\label{eq:th1}
\Delta ( C^0[T]) < \Theta
\end{equation}
where $\Theta$ is some predefined threshold. Using the distance measure, we sort all the available text samples $T' \in \T_L + \T_U$ based on their similarity to $T$ and pick the first $N$ nearest neighbors $T'_{(1)}, \ldots , T'_{(N)}$.

We can now calculate the average classifier score of the sample and $1 \leq m \leq N$ of its nearest neighbors:

\begin{align}
    S_0 &= C^0[T]\\
    S_{ k } &= C^0 [T'_{(k)}] \\
    \bar{S}^m &= \frac{1}{m+1} \left( S_0 + \sum_{k=1}^{m} S_k  \right)
\end{align}

Starting from $m=1$, we check if the ambiguity is lowered enough after we averaged the scores of the $m$ neighbors, i.e.,
\begin{equation}
\label{eq:th2}
\Delta ( \bar{S}^m )>\Theta
\end{equation}

Once the new ambiguity passes the threshold, it is likely that we have improved the accuracy of the labeling. The new label for the sample $T$ will now be $\hat{L} = \mathrm{argmax}(\bar{S}^m)$. We can now add the pair of text samples with its new label $(T, \hat{L})$ to our training data and continue the process with the other high-ambiguity samples.

If \eqref{eq:th2} is not satisfied for all $m \leq N$, then the text sample $T$ will remain unlabeled and cannot be added to the training data.

\section{Experiments and results}
\label{sec:Experiments}

To verify the usefulness of the NNSI algorithm, we used two datasets taken from telephony-based commercial US-English speech conversation systems. Both datasets contain propriety and private information from businesses, customers, and users and therefore cannot be shared or made public.

The setup for our experiments included a commercial, SVM-based classifier for the intent classification. Each classifier model was trained on text and label pairs. No other parameter of this model could be tuned. 

To estimate a semantic distance, we used a pre-trained DNN-based text embedding model \cite{cer2018universal} and the cosine distance function. After tuning with held-out data, we selected the threshold $\Theta$ for \eqref{eq:th1} and \eqref{eq:th2} to be the median of ambiguity values from the unlabeled set. Consequently, half of this set is considered low-ambiguity and the other half is high-ambiguity. In addition, we set $N=10$.

\subsection{Dataset A}
\label{ssec:krg}

The first dataset was taken from a running commercial voice conversation system. Its intent classifier is required to identify 104 different intents.
The baseline labeled training set for this system, with about 3,300 samples, was created and labeled by subject matter experts. 

We collected over 200,000 anonymized audio conversations from this system over a period of several months. From these recordings, we extracted the relevant user input and automatically transcribed it using a speech-to-text system that was tuned for this conversation system.

A subset of more than 10,000 random conversations was selected for labeling. For these conversations, the speech of the main user input was manually transcribed.

The transcribed text was submitted for manual intent labeling by a group of trained workers. Each sample was labeled by at least three workers and a majority vote was applied for the final label. Samples that did not receive a majority vote were sent for additional review by experts. In addition to the 104 original intents, we added two more: “no-intent” for samples that did not have any meaningful intent and “multiple-intents” for samples that contained two or more clear intents.

From this labeled set, we randomly selected a set of 5,000 samples. Using these samples, we created two test sets: one with the STT text and the other with the manually transcribed text for the same samples. The STT set represented a real-world scenario where the input to the classifier is the STT output. The manual transcript set represented the ideal scenario we could achieve by improving the STT. The rest of the labeled samples were used as a development set.

We measured the intent classification accuracy of the baseline classifier using the labeled test sets. We tested the accuracy of the baseline classifier on the text from the STT and on the manually transcribed text. The results are listed in the first line of Table \ref{tab:ex1results}. As can be seen, the errors in the STT output cause about a 5\% degradation in the classifier's accuracy.

Our unlabeled dataset contained all the unique text samples from the STT output of the collected conversations, except for those conversations that were already included in the test sets. We did not use any manual labels or transcriptions even if they were available for samples in this set.

Once we applied the NNSI algorithm on this set, it selected and labeled 19,567 high-ambiguity samples. We added the selected samples with their NNSI labels to the original baseline training set and trained a new classifier. The error rates of the classifiers on the test set are shown on the second line of Table \ref{tab:ex1results}. As can be seen, the additional training data from NNSI improves the accuracy of the classifier in both cases.

Out of the samples selected by the NNSI algorithm, there were 615 with manual labels, that were part of the labeled development set. This allowed us to compare the labels produced by the baseline classifier and NNSI to the manual labels. We found that the selected samples have an accuracy of 67.8\%, compared to the 37.9\% average accuracy of the baseline classifier on the high-ambiguity samples set.

For additional comparison, we added two experiments. In the first, we applied the baseline classifier to the unlabeled samples and randomly selected 19,567 samples with high-ambiguity labels. We added these samples to the baseline training set with the classifier's top label and then trained a new classifier on this new training set. This was repeated 10 times for cross validation. The second experiment was similar but we selected samples with low-ambiguity labels. The results of these two experiments are also shown in Table \ref{tab:ex1results}. We can see that the NNSI algorithm out-performs both methods. Future experiments could examine combinations of NNSI-generated samples from the high-ambiguity samples and additional low-ambiguity samples.

\begin{table}[htb]
\centering
\resizebox{\linewidth}{!}{%
\begin{tabular}{@{}lllll@{}}
\toprule 
         & \multicolumn{2}{c}{STT} & \multicolumn{2}{c}{Transcription} \\
Test     & Err (\%)        & $\Delta$Err (\%)      & Err (\%)              & $\Delta$Err (\%)           \\ 
\cmidrule(l){2-3} \cmidrule(l){4-5} 
Baseline & 36.0      &      & 31.4           &           \\
\rowcolor[HTML]{EFEFEF}
\textbf{NNSI} & \textbf{33.1} & \textbf{8.2} & \textbf{28.6} & \textbf{9.0} \\
\begin{tabular}[c]{@{}l@{}}Random\\ High-ambiguity
\end{tabular} & 36.4 $\pm$ 0.1 & \color{red} -1.1 $\pm$ 0.4 & 31.9 $\pm$ 0.1 & \color{red} -1.7 $\pm$ 0.4 \\
\rowcolor[HTML]{EFEFEF} 
\begin{tabular}[c]{@{}l@{}}Random\\ Low-ambiguity
\end{tabular}  & 34.9 $\pm$ 0.1 & 3.0 $\pm$ 0.2 & 30.3 $\pm$ 0.1 & 3.5 $\pm$ 0.2  \\ \bottomrule
\end{tabular}
}
\caption{Classification errors for the different tests on the text from the STT and the manual transcription text. "Err" is the percentage of incorrect classification on the test sets and  $\Delta Err=(Err_{baseline}-Err)/Err_{baseline}$ is the relative error reduction from the baseline.}

\label{tab:ex1results}
\end{table}

\begin{figure}[htb]

\begin{minipage}[b]{1.0\linewidth}
  \centerline{\includegraphics[width=8.5cm]{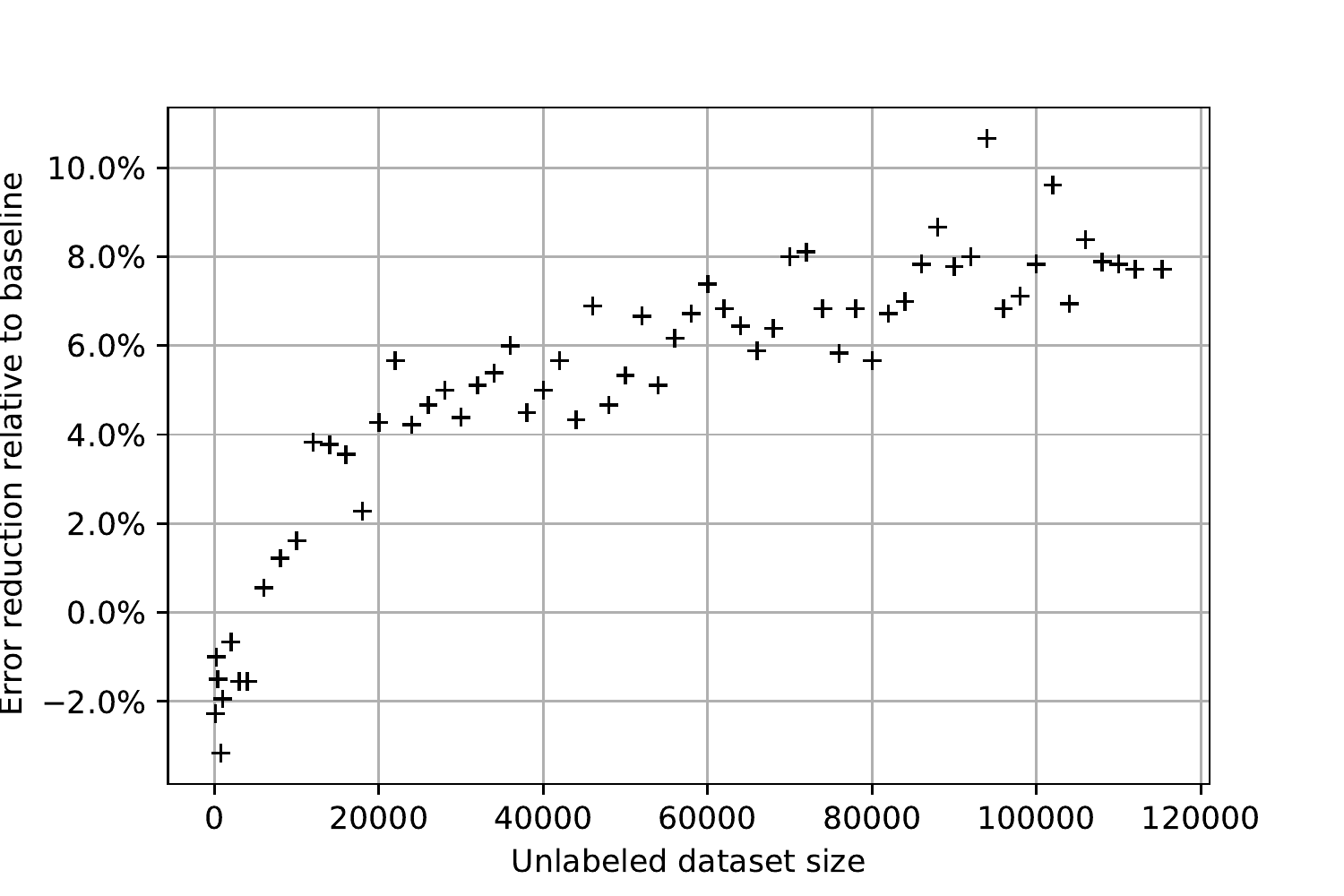}}
\end{minipage}
\caption{Error reduction relative to the baseline on the STT texts for different unlabeled dataset sizes}
\label{fig:errVsDsSizeKrg}
\end{figure}

To estimate the effect of the unlabeled dataset size, we repeated the NNSI selection and labeling procedure on random subsets of the unlabeled dataset with various sizes. The results are shown in Figure \ref{fig:errVsDsSizeKrg}. The graph clearly shows that additional data helps to improve the accuracy of the classifier.

\subsection{Dataset B}
\label{ssec:conv}
Our second dataset was composed of recordings from a call center with manual transcription and labeling \cite{convy0,convy1,convy2}. There are in total about 30,000 samples split into test (5,600), train (21,900), and development (2,500) sets.
The transcribed samples were manually labeled using 33 intents. In addition, an STT transcription was generated for all of the samples using the development set for language model adaptation.

For this dataset, we did not have any additional unlabeled samples. Instead, we split the training set into labeled and unlabeled parts. For the labeled part, we kept the labeling and used the manually transcribed text because we wanted to match the base training set created by the system developer (without STT errors). This part was used in training the baseline classifier. For the unlabeled part, we used the rest of the samples without labeling and used only the STT text sample, similar to what would happen in a real-world scenario.

We experimented with different sizes for the baseline training part, from $0.5\%$ to $25\%$ of the data. For each training data size, we performed a 10-fold cross-validation with different random splits.

As in the previous experiment in Section \ref{ssec:krg}, we created two test sets out of the test samples: one with the manually transcribed text and the other with the text from the STT.

\begin{figure}[htb]

\begin{minipage}[b]{1.0\linewidth}
  \centerline{\includegraphics[width=8.5cm]{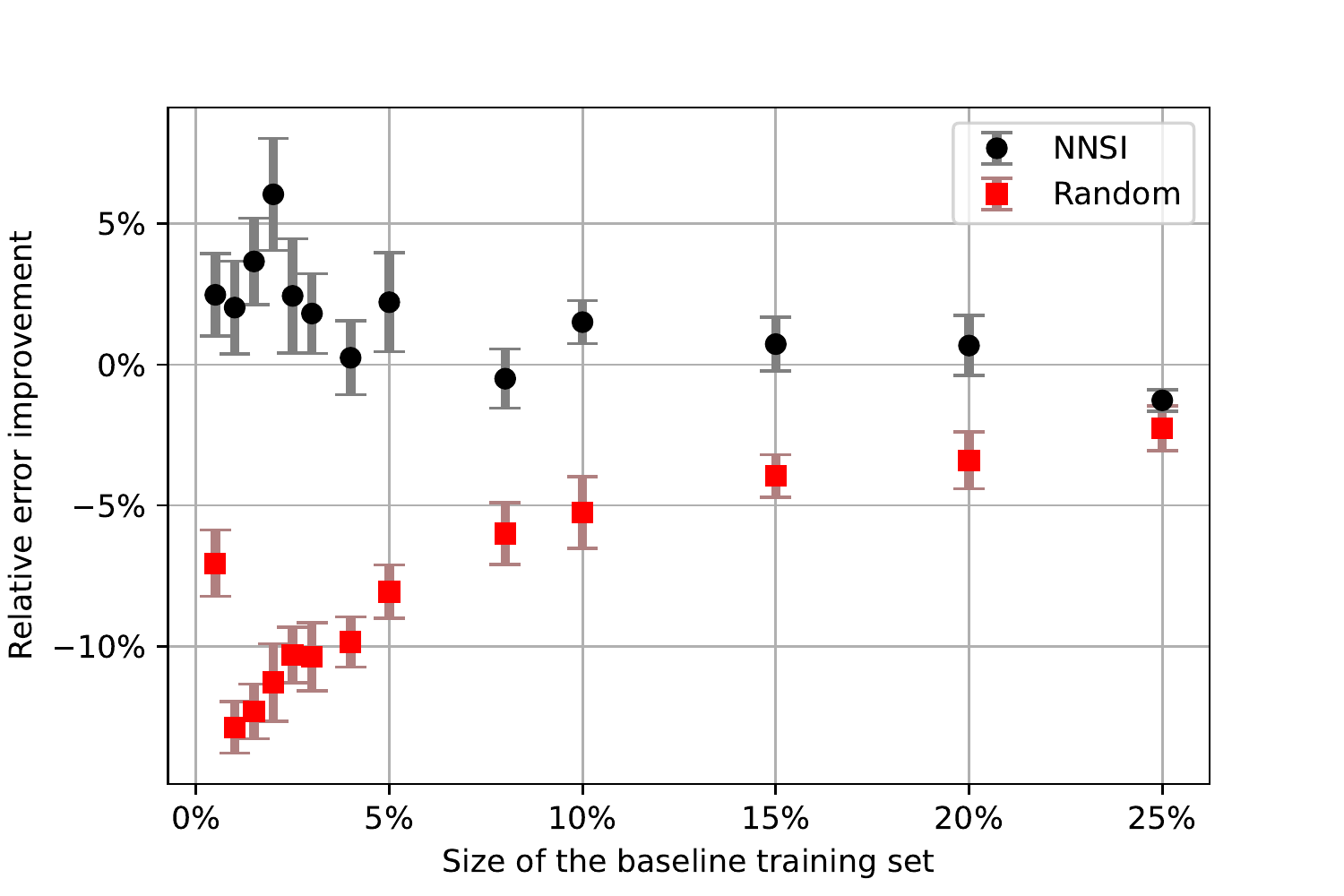}}
    \centerline{(a) Tests on manual transcription text}\medskip
\end{minipage}
\begin{minipage}[b]{1.0\linewidth}
  \centerline{\includegraphics[width=8.5cm]{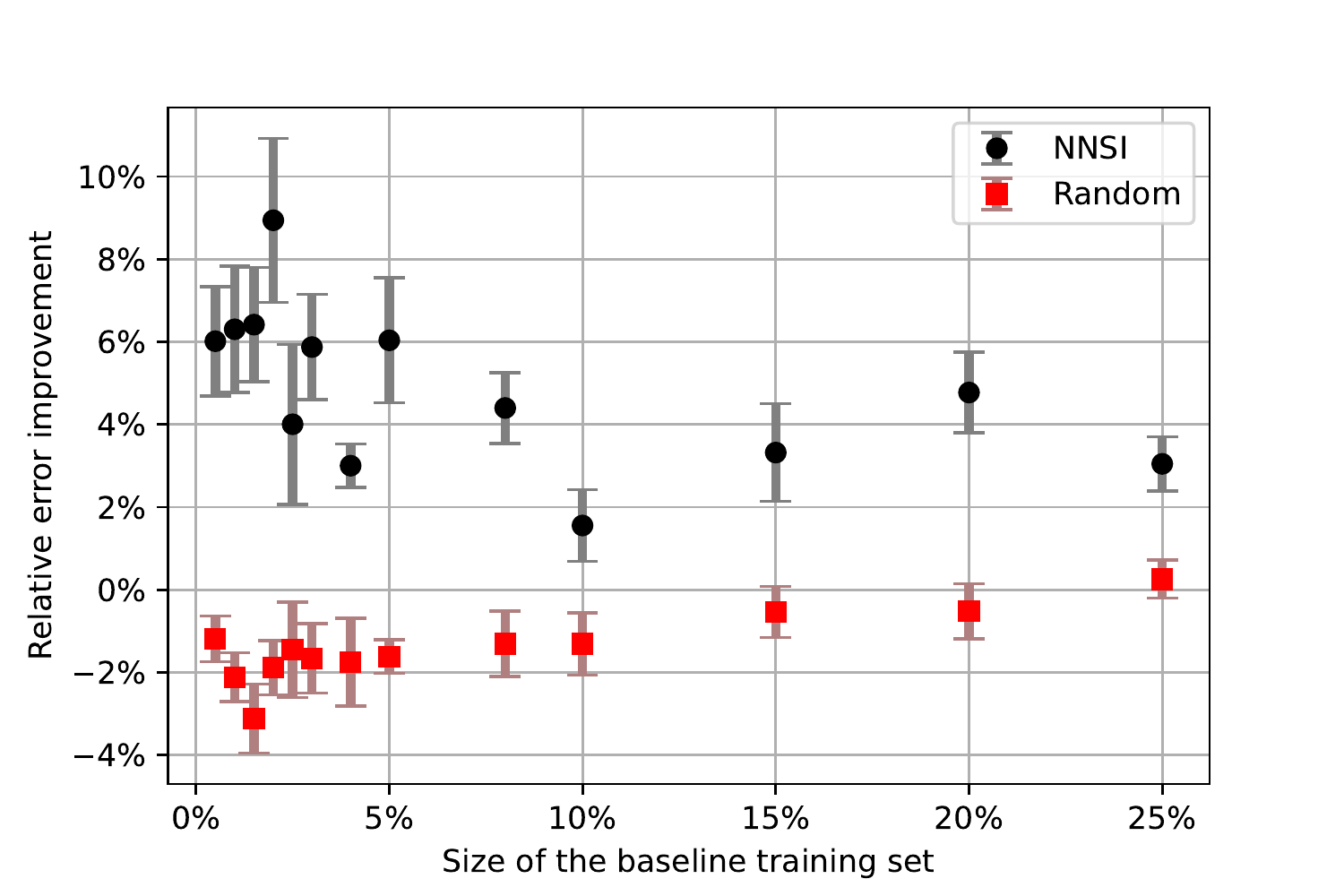}}
    \centerline{(b) Tests on STT text}\medskip
\end{minipage}
\caption{Error reduction relative to the baseline for different splits of the training set.
Error bars represent the standard error range.}
\label{fig:errVsDsSizeConv}
\end{figure}

Figure \ref{fig:errVsDsSizeConv} shows the relative classification error improvement when using NNSI to add samples from the unlabeled set to the baseline training set. For comparison, in each one of these experiments, we randomly selected the same number of high-ambiguity samples from the unlabelled part and added them to the training using the labels from the baseline classifier (not the original manual labels). Figures \ref{fig:errVsDsSizeConv} (a) and (b) show the results for the manually transcribed text and the STT text, respectively.

In both cases, NNSI is most beneficial when the baseline labeled set is small and hence the number of unlabeled samples is much greater than the baseline set. In these cases, the random set performs much worse because of the low accuracy of the labels generated by the baseline classifier.
We saw a significantly larger improvement for the STT test set in Figure \ref{fig:errVsDsSizeConv} (b), which represents the real-world scenario. This is notable even for larger split ratios. This probably happens because the unlabeled STT data helps to bridge the gap between the training data (clean transcript) and the test data (STT with errors).

\section{Conclusion}
\label{sec:conclusion}
We presented NNSI, a novel semi-supervised learning algorithm.
The NNSI algorithm can help improve a classifier's accuracy by selecting and labeling additional training samples from a large set of unlabeled data. Our experiments on  two real-life datasets and showed that it can produce classifiers with improved accuracy at various working points. A particularly interesting finding is that the algorithm is robust and performs even better with noisy data coming out of the speech-to-text system.
Additional work is needed to further improve the algorithm by using different graphs, averaging methods, or different similarity measures, and by testing the impact on other types of classifiers or data types. Another possible direction for research on spoken conversation use cases is to explore whether this method can be applied for end-to-end intent classification directly from speech.

\section{Acknowledgements}
\label{sec:ack}
The authors of this paper would like to thank the IBM Global Business Services team that helped us with the data collection and the IBM Research team led by Mohamed Nasr for performing the data labeling.


\bibliographystyle{IEEEbib}
\bibliography{strings,refs}

\end{document}